\documentclass{llncs}
\usepackage{times}
\usepackage{epsfig}
\usepackage{subcaption}
\usepackage{graphicx}
\usepackage{amsmath}
\usepackage{amssymb}
\usepackage{bbm}
\usepackage{copyrightbox}
\begin{document}
\title{Learning Self-Supervised Audio-Visual Representations for Sound Recommendations}
\author{Sudha Krishnamurthy}
\institute{Sony Interactive Entertainment, San Mateo, CA \\
	      \email{\tt krishnadots@gmail.com}
       }

\maketitle

%%%%%%%%% ABSTRACT
\begin{abstract}
We propose a novel self-supervised approach for learning audio and visual representations from unlabeled videos, based on their correspondence. The approach uses an attention mechanism to learn the relative importance of convolutional features extracted at different resolutions from the audio and visual streams and  uses the  attention features to encode the audio and visual input based on their correspondence. We evaluated the representations learned by the model to classify audio-visual correlation as well as to recommend sound effects for visual scenes.  Our results show that the representations generated by the attention model improves the correlation accuracy compared  to the baseline,  by 18\% and the recommendation accuracy by 10\% for VGG-Sound, which is a public video dataset. Additionally,  audio-visual representations learned by training the attention model with cross-modal contrastive learning further improves the recommendation performance, based on our evaluation using VGG-Sound and a more challenging dataset consisting of gameplay video recordings.
\end{abstract}

\section{Introduction}
\begin{figure}[t]
		\centering
		\includegraphics[height=0.6in]{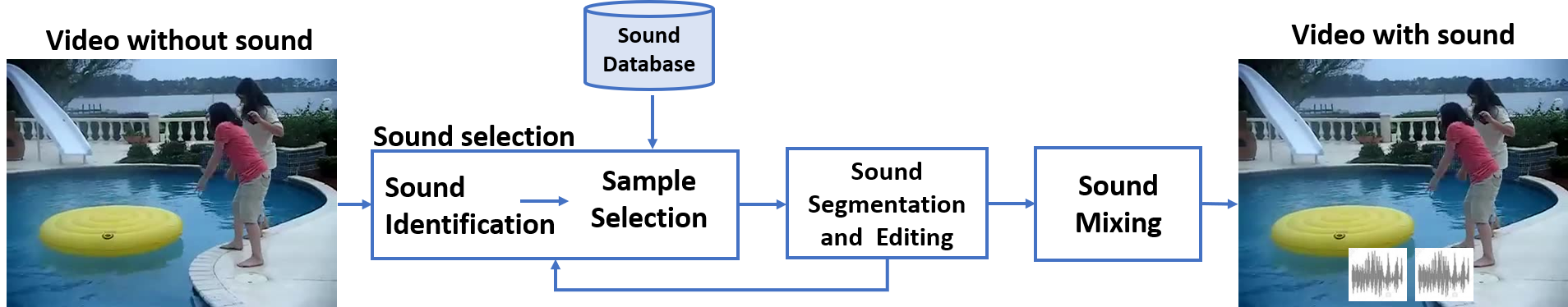}
		\caption{Sound matching and mixing for videos (images best viewed in color)}
		\label{fig:foley}
\end{figure}

Learning audio and visual representations based on their correspondence, by training on videos without explicit annotations for the objects and sounds that appear in the videos, is a challenging problem. The learned representations are useful for some of the audio-visual (A-V) content creation tasks. Our motivation is to use these representations to recommend sound effects based on the visual scene, in order to assist sound designers in creating entertaining video content, such as short trailers or video games.  Currently, sound effect generation (a technique called Foley)  for video game or movie content generation is a time-consuming and  iterative process, as shown in Figure~\ref{fig:foley}. Given a silent video sequence, in the first step the sound designer identifies the relevant sound categories by understanding the visual scene. For example, for the video frame shown in Figure~\ref{fig:foley}, a designer  may identify water-related sounds and human voices as sounds that correspond to the visible objects. Additionally, ambient sounds, such as wind related sounds, may also be relevant, even though this background context may not be visible. Next, specific sound samples for the selected sound categories are retrieved from a  sound database. This sound selection and retrieval process is iterative, since there may be numerous  variations of sound samples  within a category. The  selected sound samples are then used to create the desired sound effects for the video. Our goal is to train a neural network model to learn audio and visual representations based on their correspondence  and use those representations to  assist in the sound selection step, by recommending the top matching sound samples (instances) as well as their respective categories from a given audio list or  database, for a visual scene.  The recommended sound effects may either correspond to objects directly visible in the scene or may be ambient sounds corresponding  to the background context.

Videos in-the-wild provide rich information for learning audio-visual correspondence because of the diversity of objects and sound sources that appear in them. However, it is challenging to get object and sound labels for video datasets with scenes containing an  unconstrained number of objects and  mixed with noisy audio containing multiple sound sources. Moreover, our goal is to recommend not just the best matching high-level sound categories, but also specific sound instances within those categories, and getting instance-level sound labels for videos, in order to train supervised models is hard. Hence, we propose self-supervised approaches that  use the inherent temporal alignment of sounds with video frames in video datasets as implicit labels for learning audio-visual representations. Our self-supervised model uses a two-stream deep convolutional network consisting of a novel attention-based encoder and a projection module, that is trained on audio-visual pairs that are either correlated or uncorrelated. The model is trained to project correlated audio and visual representations closer in the latent space compared to uncorrelated representations.
The attention encoder generates  audio and visual representations based on the relative importance of local and  global convolutional features extracted at different resolutions from the audio and visual streams \cite{jet_iclr2018}. The attention mechanism helps in disentangling the latent features extracted from videos having multiple objects and sound sources, without using object detection or audio source separation. Additionally, we trained the attention model to learn audio-visual  representations using cross-modal contrastive learning without relying on data augmentation, unlike  previous unimodal contrastive learning methods \cite{chen_icml2020_simclr,oord_2019} that rely on data augmentations. 

We evaluated the learned representations by using them to (1) classify audio-visual correlation, and  (2) recommend sound samples corresponding to visual scenes.  Our experimental results show that the audio and visual  representations generated by the attention model improves the classification accuracy by 18\% and the recommendation accuracy by 10\% for VGG-Sound \cite{chen_icassp2020}, which is a public dataset consisting of videos in-the-wild. Additionally, training the attention model with cross-modal contrastive learning results in better representations and improves the recommendation performance further, even without data augmentation, based on our evaluation on VGG-Sound and a more challenging and novel video dataset that we created by recording gameplay sessions of people playing different video games.

\section{Related Work}
\label{sec:rel}
%single source audio
Self-supervised approaches that have been previously proposed to learn  audio-visual correspondence and representations, typically  assume that  the  videos either have a single dominant audio signal and visual theme, or a small but fixed number of  visual objects and audio sources. For example,  the approach in \cite{owe_cvpr2016}	uses a recurrent network to learn A-V correlation from unlabeled videos having a single dominant sound that is generated by a physical impact. The model in \cite{zhao_eccv2018_pixel} learns A-V correspondence from unlabeled videos whose dominant sound is a musical recording, by disentangling and correlating audio mixtures from a pair of videos. In \cite{arand_iccv2017_L3,arand_eccv2018}, a convolutional two-stream network is trained to classify the A-V correspondence  based on cross-modal self-supervision using videos that also have a single dominant audio source. Unlike the above approaches, the  multi-instance multi-label (MIML) framework in  \cite{gao_eccv2018_miml} learns from videos with multiple but a small and  fixed number of objects and audio sources. It uses a pretrained image classifier to detect a fixed number of objects, then uses this visual guidance to separate the corresponding noisy audio  into a fixed number of audio sources, and learns the correlation between the detected objects and separated  audio sources.  An attention mechanism is used in \cite{seno_cvpr2018}, to localize sound sources in visual scenes and learn A-V representations. In  \cite{owens_eccv2018_multisense}, self-supervised A-V representations are learned by predicting the temporal alignment between audio and visual frames.

Our approach also learns audio-visual representations and correlation from unlabeled videos, but does not make any assumptions about the number of visual objects or audio sources in the video. The training videos can be in-the-wild and the learned A-V correspondence may include sounds that are not just related to the visible objects, but also to the ambient context, such as sound of wind blowing. 
%with no explicitly visible source,  
Instead of performing  audio source separation or object detection, we use an attention-enhanced model to focus on relevant local and global features extracted at different resolutions  from the audio-visual streams, to encode the audio and visual representations based on their correspondence. We also propose the use of cross-modal contrastive approach for learning self-supervised audio and visual representations based on their correlation.

\section{Datasets}
\label{sec:dataset}
We used the following video datasets to train  and evaluate our self-supervised audio-visual models:

1) {\bf{VGG-Sound dataset}} \cite{chen_icassp2020}: This dataset provides a list of short audio clips corresponding to "in-the-wild" videos posted online by users.  We were able to download 5,829 videos from the training split and 428 videos from the test split. Each video clip has a  label that indicates the dominant sound or visual theme in the video. However, in some cases these labels are noisy and for some of the videos the sound  does not correspond to any visible object in the  clip, making it challenging to infer the correlation.  After preprocessing, there were  305 unique sound labels in the training set.  These sound labels were used only for evaluating the models and not for training. 

2) {\bf{Gameplay dataset}}: As mentioned earlier, one of our motivations for learning audio-visual correlations is to automatically recommend sound effects corresponding to visual scenes, which can be used to produce entertainment videos, such as video games. Hence, we created a gameplay dataset by recording videos of the  gameplay sessions of some video gamers. These videos span different video game genres and most of the visual scenes include complex backgrounds and multiple objects. Unlike the VGG-Sound dataset, which largely comprises videos with a single dominating sound, each gameplay video consists of a noisy monaural audio mixture composed of an unknown number of sound sources, making it hard to separate the noisy mixture into its clean audio sources. 
Furthermore, there are no annotations for any of the visual objects or sounds that appear in the videos, which makes this dataset considerably more challenging for training and testing compared to the VGG-Sound dataset.

We extracted video frames from each of the videos in the above datasets at 1 fps using {\em ffmpeg}. We also extracted the  audio  from each of the videos and segmented it into 1-second audio samples. For each 1-second audio sample,  a 64x100-dim log-mel-spectrogram audio feature  was  generated for training by computing the logarithm on the mel-spectrogram feature extracted for the sample at a sampling rate of 48 KHz, using the {\em {librosa}} library \cite{librosa_2015}. We trained the self-supervised correlation models separately on the two datasets with audio-visual input pairs that are either correlated or uncorrelated. Each input pair consists of a  video frame  and the log-mel-spectrogram feature extracted from a 1-sec audio sample and if both are temporally aligned in a given video, then they are correlated (positive sample). To create  uncorrelated pairs (negative samples) for training, we used slightly different methods for the two datasets. For the Gameplay dataset, which has longer videos  with varying audio, an uncorrelated  pair consists of  a video frame paired with a 1-second  audio feature extracted either from a different video or from  a  different timeframe of the same video. To create a negative pair for VGG-Sound, which has shorter video clips, we paired a video frame with a 1-second audio feature extracted from a video with a different sound label.

\section{Self-Supervised Audio-Visual Representation Learning}
\label{sec:models}
Given an input batch of video frames paired with audio, we trained deep convolutional network models to learn audio and visual embeddings (representations) and used the embeddings for downstream tasks, such as recommending sounds relevant to a visual scene. The model has 3 components: {\em encoder, projector, and a loss function}. The backbone of the network is a two-stream audio-visual encoder, $E=(E_v, E_a)$, which generates the visual embedding $E_v(v_i)$, and audio embedding $E_a(a_i)$ for an input pair ($v_i, a_i$) consisting of an image (video frame) $v_i$ and audio input $a_i$. The projector, $P$, transforms the generated embeddings into suitable inputs for computing the loss function.  
%then projects these generated  audio and visual embeddings  onto a common embedding space and in some cases, may lower the dimension of the embeddings. 
The encoder and projector are trained together end-to-end by a loss function, $L$. The goal of training is to learn embeddings such  that correlated audio and visual inputs are nearby in the latent space, while uncorrelated audio and visual embeddings are farther apart. After the network is trained, we only use the audio-visual encoder for downstream tasks. In this section, we describe the different architectural choices for the encoder and projector, as well as the loss functions that we experimented with and in Section~\ref{sec:exp}, we compare their performance  on downstream tasks.

\begin{figure*}[t!]
	%\centering
	\begin{subfigure}[t]{0.3\textwidth}
		\centering
		\includegraphics[height=2.0in]{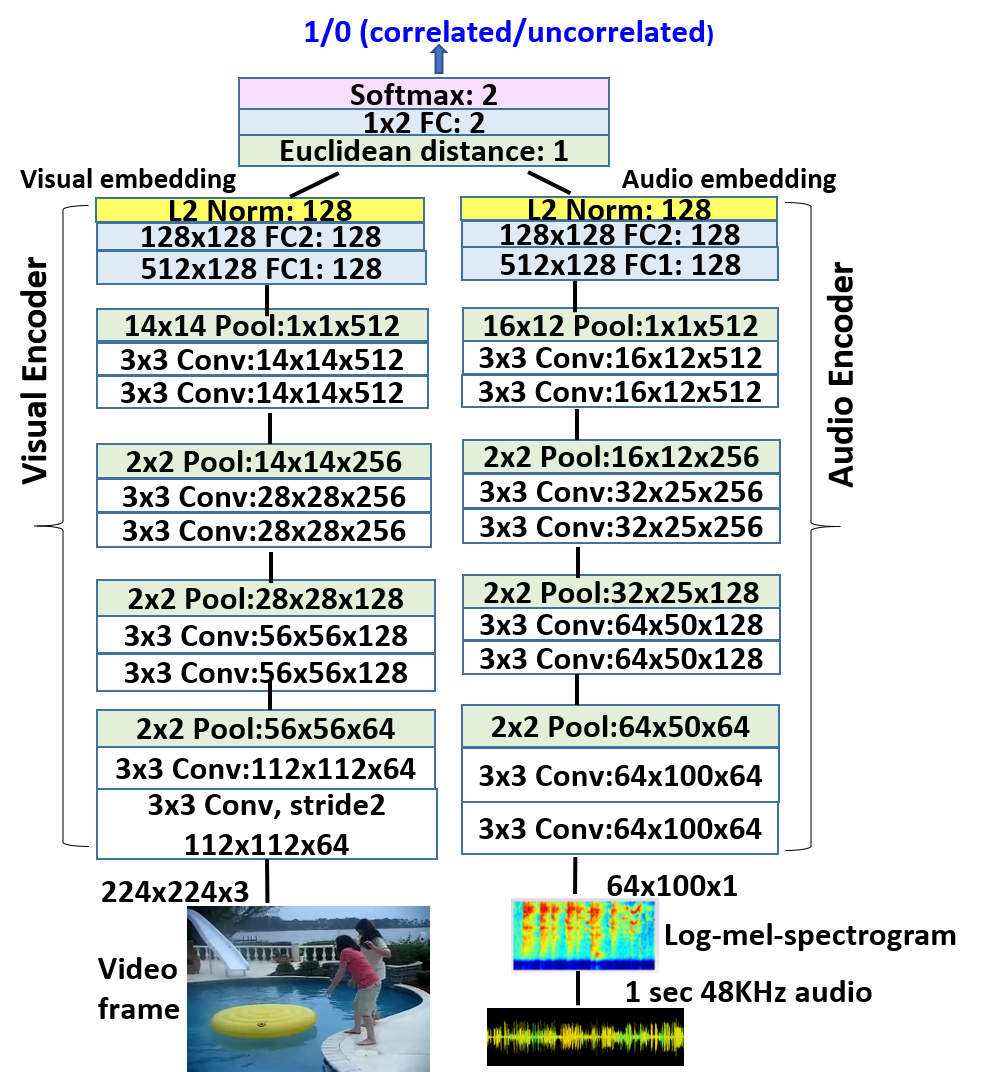}
		\caption{Baseline 2-stream A-V network}
		\label{fig:ots}
	\end{subfigure}
	\hfill
	\begin{subfigure}[t]{0.6\textwidth}
		%\centering
		\includegraphics[height=2.0in]{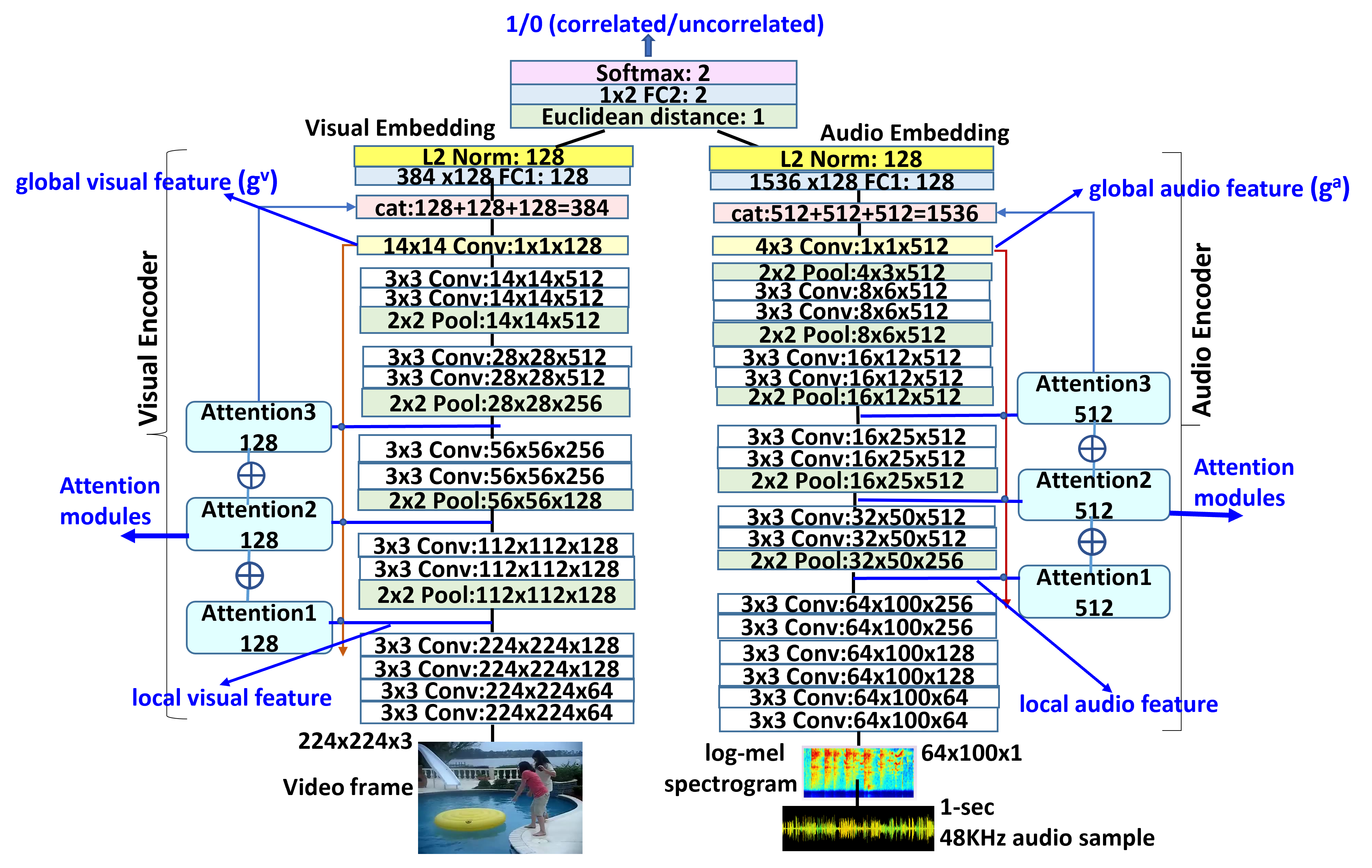}
		\caption{2-stream A-V network with \\attention-based encoder}
		\label{fig:att}
	\end{subfigure}
	\caption{Audio-visual correlation networks. Each layer shows the type of the layer and the  dimensions of the output.}
\end{figure*}

\subsection{Baseline Model}
\label{sec:av_base}
Our baseline model  is shown in Figure~\ref{fig:ots} and it is a slight variation of the audio-visual correlation model presented in \cite{arand_eccv2018}. 
The encoder network $E$ in the baseline model uses a two-stream network  consisting of a series of convolutional blocks followed by a few fully-connected layers for encoding the visual and  audio inputs. Each convolutional block consists of two convolutional ({\em{conv}}) and one max-pooling layers.  Batch normalization followed by ReLU nonlinearity is applied to the output of each {\em{conv}} layer. 

The input to the visual stream is a 224x224-dim RGB video frame. The input to the audio stream  is a 64x100-dim log-mel-spectrogram feature extracted from a 1-second audio sample, as described in Section~\ref{sec:dataset}. This audio feature is processed as a grayscale image by the convolutional audio encoder. At the end of the visual and audio encoding pipeline, an $L2$-normalized 128-dim embedding is generated for the visual and audio input, respectively. These audio and visual embeddings are then input to the projection subnetwork $P$, which uses a fully-connected layer (FC in Figure~\ref{fig:ots}) to compute the correlation score as a function of the Euclidean distance between the  embeddings and the bias parameter learned by this layer  represents  the distance threshold that separates the correlated and uncorrelated pairs \cite{arand_eccv2018}.  The correlation score is used by the binary classification ({\em{softmax}}) layer to predict if the audio-visual input pair is correlated or not. The baseline model is trained with batches containing an equal number of  correlated and uncorrelated audio-visual pairs extracted from the datasets described in Section~\ref{sec:dataset}.  The network is trained to minimize the binary cross-entropy loss and in the process learns to generate audio-visual representations  that are closer in the latent space when they are correlated and farther when they are uncorrelated.

\subsection{Attention-based Model}
\label{sec:av_att}
The baseline model learns a coarse-grained correlation between the image and audio input. However, as mentioned in Section~\ref{sec:dataset},  the visual scene in a video frame may depict multiple objects and events. Similarly, the noisy audio input may be a mixture of different sound sources, some of them more dominant than the others. Hence, an encoder that weighs the influence of features from different input regions based on their relative importance can learn better encoded representations. To enable this, we enhance the audio and visual convolutional encoders  with trainable soft attention layers \cite{jet_iclr2018} and combine the encoder with the  same projection module  used in the baseline model.  This attention-based audio-visual correlation network is shown in Figure~\ref{fig:att}.

In the baseline model in Figure~\ref{fig:ots}, only the audio and visual embeddings generated at the end of the encoding pipeline were used  for learning the A-V correlation. However, in the attention model, we extracted both global and local features representing different audio and visual concepts to generate the audio-visual embeddings and  learn the correlation based on that. The local features were extracted at different spatial resolutions from three  different intermediate layers of the audio and visual encoder pipelines, as shown in Figure~\ref{fig:att}. Let $L^a=(l_1^a, l_2^a, \ldots, l_n^a)$ and  $L^v=(l_1^v, l_2^v, \ldots, l_n^v)$  denote the set of local features extracted from $n$ intermediate layers (in our model $n=3$) of the audio and visual streams, respectively. Each local feature  was extracted after a convolutional block but before the  max-pooling layer of that block (local features extracted after the pooling layer lowered the spatial resolution did not result in good performance). Also, let   the output of the final convolutional layer of the audio and  visual encoding pipelines be denoted as $g^a$ and $g^v$,  which we consider as the global audio and visual feature, respectively (see Figure~\ref{fig:att}). We now explain how the attention-based visual encoder  uses the local and global features to generate the visual embeddings (a similar procedure was used to generate the audio embeddings).   

1) $w_i^v = conv(l_i^v + g^v), i=1,\ldots,n$:  First, the attention weight $w_i^v$ of each local feature  $l_i^v$ is computed with respect to the global feature  $g^v$  by adding the two features and passing the sum through a 1x1 convolution operation. Intuitively, these attention weights represent the compatibility between the local and global features. The dot product is an alternative way to compute this compatibility score. If the dimensions of $l_i^v$ and  $g^v$ are different, then $l_i^v$ is projected to match the dimension of $g^v$ before the two features are added.  

2) $g_i^v = A_i^v*l_i^v, i=1,\ldots,n$: The attention weights (from Step 1) are normalized through a softmax operation to yield the normalized weights, $A_i^v$. Then each local feature $l_i^v$, is weighted through element-wise multiplication with its normalized attention weight $A_i^v$, to yield the corresponding  attention feature $g_i^v$.

3) $g_{att}^v = g_1^v \oplus g_2^v \oplus \ldots \oplus g_n^v$: The resulting $n$ attention features  are then concatenated to yield the global attention feature $g_{att}^v$ for the visual stream.

4) $e^v = norm(project(g_{att}^v))$:  Finally,  $g_{att}^v$ is projected to a lower dimension (we used 128-dim) and normalized using $L2$-norm to yield the visual embedding $e^v$. 
The audio encoder uses the same sequence of operations to generate the 128-dim audio embedding, $e^a$, for an audio sample. 

The 128-dim  visual and audio  representations, $e^v$ and $e^a$, generated in Step 4 by the attention-based visual and audio encoders  are then input to the projection module, which computes the correlation based on the distance between these representations, as described in Section~\ref{sec:av_base}. This attention model was trained for audio-visual correlation using binary cross entropy loss and  the same input as the baseline model.  In addition, it was trained with a combination of binary cross entropy ($L_{bce}$) and margin-based contrastive loss ($L_{m}$), $L=L_{bce} + L_{m}$. Given the visual and audio representations, $e^v$ and $e^a$, and the ground-truth correlation label, $y \in \{0, 1\}$, indicating if the input audio-visual pair is correlated ($y=1$) or not ($y=0$), $L_{m}$ is defined as follows \cite{had_2006}:
\begin{equation}
\label{eqn:margin}
%\footnotesize
L_{m} (e^v, e^a, y) = (y)  dist(e^v, e^a) + (1-y) max(0, m-dist(e^v, e^a))
\end{equation}
where $m$ is a hyperparameter denoting the margin separating the correlated and uncorrelated A-V pairs and $dist$ measures the Euclidean distance between the audio and visual embeddings. The label $y$ is obtained in a self-supervised manner based on the temporal alignment between the video frame and audio sample, as described in Section~\ref{sec:dataset}. We describe the training hyperparameters and the performance using different loss functions in Section~\ref{sec:exp}.

\subsection{Unlabeled Contrastive Learning}
\label{sec:av_contrast}
The two models described above are trained to classify binary audio-visual correlation using balanced batches containing an equal number of correlated and uncorrelated input pairs. In the process, the models  learn self-supervised audio-visual representations that are closer in the latent space if they are  correlated and farther if they are  uncorrelated, which are then used for different downstream tasks. In this section, we present another approach that uses contrastive learning to train  a model directly for self-supervised representation learning, without the need for a classification objective. When recommending audio samples from a large database, the fraction of uncorrelated (negative) audio samples is much higher than the fraction of positive samples that correspond to a given visual input. To reflect a similar distribution during training and learn better representations, this third approach compares each visual input with a larger population of negative audio samples, without increasing the batch size.   

Recently, contrastive learning methods have been used for self-supervised visual representation learning \cite{chen_icml2020_simclr,chen_2020_moco2,grill_2020_byol,he_cvpr2020_moco}. These methods  minimize the distance between representations of positive pairs comprising augmented views of the same image, while maximizing the distance between representations of negative pairs comprising augmented views of different images. Some of these methods rely on large batch sizes \cite{chen_icml2020_simclr,chen_nips2020_simclr}, large memory banks \cite{chen_2020_moco2,he_cvpr2020_moco,misra_cvpr2020_pirl}, or  careful negative mining strategies. While these  approaches rely on data augmentation to learn self-supervised unimodal representations, we  propose a cross-modal contrastive learning approach for learning both audio and visual representations, without relying on data augmentation. 

Our contrastive network architecture uses the same attention-based 2-stream audio-visual encoder shown in Figure~\ref{fig:att} as its backbone, but uses a different projection module, training strategy, and  loss function. We experimented with two different projection architectures: a single-layer linear MLP and a 2-layer MLP with a ReLU non-linearity between the 2 layers. The projector for contrastive learning does not compute the distance between the audio-visual embeddings output by the encoder, to classify the audio-visual correlation, as was done in the previous two  models. Instead, the projected embeddings are used to compute the contrastive loss \cite{chen_icml2020_simclr}. Previous contrastive methods  introduced two  loss functions for unimodal  representation learning: (a)  normalized temperature-scaled cross-entropy (NT-Xent) loss function \cite{chen_icml2020_simclr} and (b) InfoNCE loss function \cite{oord_2019}, which is based on noise contrastive estimation.  We adapted these loss functions to train our model end-to-end for cross-modal representation learning, as follows.

The input batch used to train the baseline and attention models in the previous subsections includes an equal number of positive and negative audio-visual pairs. In contrast,  the training batch for contrastive learning models consists of $N$ audio-visual pairs, denoted by $X=\{(v_1,a_1), (v_2,a_2),\ldots,(v_N,a_N)\}$, where each visual input $v_i$ is correlated with audio input $a_i$, but is not correlated with audio input $a_k, k\neq i$. In other words,  each $v_i$ is paired with one positive audio sample, while the remaining uncorrelated $N-1$ audio samples in the batch serve as negative audio samples for $v_i$, which obviates the need to explicitly  sample negative pairs for training. This allows the model to learn from a larger population of negative pairs without increasing the batch size.   The NT-Xent loss function for a positive audio-visual pair ($v_i, a_i$) is defined in Equation~\ref{eqn:ntxent}. The training objective is to minimize the average loss computed across all the positive audio-visual pairs.
%\hspace{-0.2in}
\begin{equation}
\label{eqn:ntxent}
%\footnotesize
%\small
L_{nx}(v_i, a_i) = -log \frac{exp(sim(z_{v_i}, z_{a_i})/\tau)}{\sum_{k=1}^{N} \mathbbm{1}_{[k \neq i]} exp(sim(z_{v_i}, z_{a_k})/\tau)}
\end{equation}

%\hspace{-0.2in}
where   the embeddings that are output by the A-V encoders ($E_v$  and $E_a$) for a visual input $v_i$, and an audio input $a_k$, are projected by the  projection module $P$ to generate $z_{v_i}=P(E_v(v_i))$ and $z_{a_k}=P(E_a(a_k))$, respectively; $sim$ is a similarity function (we used cosine similarity) that computes the similarity between the projected visual and audio representations;  $\tau$ is the temperature hyperparameter used for scaling the similarity values; $\mathbbm{1}_{[k \neq i]} \in \{0,1\}$ is an indicator function that evaluates to 1 iff  $k \neq i$.

We also adapted the InfoNCE loss function to train the contrastive learning model for audio-visual representation learning.   The  InfoNCE loss function for self-supervised unimodal representation learning was formulated in \cite{oord_2019}. Given a batch $X$ of $N$ audio-visual pairs as described above, where  each image $v_i$ in the batch is paired with one positive audio sample $a_i$ while the remaining $N-1$ audio samples serve as negative samples for that image, the InfoNCE loss function computes the categorical cross-entropy of correctly classifying the index of the positive audio sample for each image in the batch.  To compute this loss, we first computed the similarity between each image $v_i$ and audio sample $a_k$ in a batch as $f(v_i, a_k) = sim(z_{v_i}, z_{a_k})/\tau$, by using cosine similarity and scaling it by the temperature  $\tau$. As in Equation~\ref{eqn:ntxent},  $z_{v_i}$ and $z_{a_k}$ are the projected representations 
%by the projection module 
for $v_i$ and $a_k$, respectively. These scaled similarity values then serve as the logit values for computing the categorical cross-entropy loss.  While the denominator term in  Equation~\ref{eqn:ntxent} includes only the similarity values for uncorrelated pairs, the InfoNCE loss computation includes the similarity values for correlated and uncorrelated pairs in the denominator.

%Gameplay 0.7
\begin{table*}[t]	
	\caption{Performance evaluation on VGG-Sound. The encoder, projector, and loss function for each model are described in Section~\ref{sec:models}.}	
	\parbox{0.3\textwidth}{
		
		\resizebox{0.3\textwidth}{!}{
			%\centering
			\begin{tabular}{||l|l|c||}
				\hline \hline
				{\bf{Encoder}} & {\bf{Loss}} & {\bf{Acc(\%)}}  \\
				\hline  
				baseline&BCE & 69.5   \\
				\hline
				attention&BCE+margin& 87.4   \\
				&contrastive & \\
				\hline
				fine-tuned&BCE+margin& 87.8   \\
				attention& contrastive& \\
				%{\bf{self-att+siamese}} & 87.3\\
				\hline  
			\end{tabular}
		}
		\subcaption{A-V correlation accuracy for VGG-Sound dataset}		
		\label{tab:cor}	
	}
	\hspace{0.02in}
	\parbox{0.7\textwidth}{		
		%\centering	
		\resizebox{0.7\textwidth}{!}{
			\begin{tabular}{||c|l|l|l|c|c||}
				\hline \hline
				{\bf{ID}}&{\bf{Encoder}} & {\bf{Projector}} & {\bf{Loss function}}& {\bf{Category-level acc (\%)}} & {\bf{Sample-level acc (\%)}} \\
				%&&&& {\bf{acc (\%)}}&{\bf{acc (\%)}}\\
				\hline  
				1&baseline&linear&BCE& 14.1 & 8.6\\
				2&attention&linear&BCE&15.6 &13.3\\
				3&attention&linear&BCE+margin-contrastive&24.2 &18.8\\
				\hline
				&{{Contrastive}}&{{Learning}}&{{Models}}&&\\
				\hline
				4&attention&nonlinear &InfoNCE& 28.1 & 23.4 \\		
				5&attention&linear & InfoNCE &{\bf{38.3}} & {\bf{31.3}}\\
				%6&attention(500 epochs)&linear & InfoNCE &35.9 & {\bf{33.6}}\\
				6&attention&nonlinear &NT-Xent& 32.0 &27.3 \\
				7&attention&linear & NT-Xent &33.6 & 28.9 \\		   
				\hline  
			\end{tabular}
		}
		\subcaption{Top-10 recommendation accuracy for VGG-Sound dataset}
		\label{tab:rec_vgg}
	}	
\end{table*}  

\section{Performance Evaluation}
\label{sec:exp}
In this section, we compare the performance of the models presented in Section~\ref{sec:models} on two tasks: 1) classifying the  audio-visual correlation and 2) recommending sound samples relevant to a visual scene. All of the models  were trained using the Stochastic Gradient Descent (SGD) optimizer with Nesterov momentum, with a momentum value of 0.9 and weight decay of $10^{-4}$. The SGD optimizer achieved better generalization than the Adam optimizer.  The models were trained and evaluated using the VGG-Sound and Gameplay datasets described in Section~\ref{sec:dataset}. The initial learning rate was selected by running a grid search on values in the range $[1 \times 10^{-4}, 1.0]$ and then adaptively decayed during training. When training with VGG-Sound, the best initial learning rate was 0.01, whereas for the larger Gameplay dataset, it was 0.001.   We trained all of the models on dual Nvidia 2080Ti GPUs. The baseline model was trained with a batch size of 128, whereas the attention-based models which have  a larger number of parameters were trained with a batch size of  32. The training was regularized using early stopping with the training terminated if there was no improvement in the validation performance for 20 consecutive epochs. To train  models with margin-based contrastive loss (Equation~\ref{eqn:margin}), we  chose margin $m=0.1$ based on  grid search. Similarly, to train the models with InfoNCE and NT-Xent losses (Section~\ref{sec:av_contrast}), we chose $\tau=0.5$ for the temperature hyperparameter, based on grid search.

To train the baseline and attention models for  A-V correlation (Section~\ref{sec:av_base} and ~\ref{sec:av_att}),  we used a balanced training dataset containing an equal number of correlated and uncorrelated A-V pairs, by randomly selecting approximately the same number of  video frames from each of the videos and pairing them with a 1-second audio sample. This resulted in a balanced training set of 58,290 A-V pairs and validation set of 1280 pairs for VGG-Sound. For the Gameplay dataset, the balanced training set contains 214,168 A-V pairs and validation set contains 35,562 pairs. The contrastive learning models for directly learning audio and visual representations  were trained on 5,829 correlated A-V pairs created by pairing one randomly extracted video frame from each VGG-Sound training video with its corresponding 1-second audio sample. As mentioned in Section~\ref{sec:av_contrast},  uncorrelated (negative) A-V pairs are generated implicitly from within a  batch when training contrastive models. No data augmentation was used for any of the models.

\begin{figure*}[ht]	
	\begin{subfigure}[t]{\textwidth}
		\centering
		\includegraphics[height=1.1in]{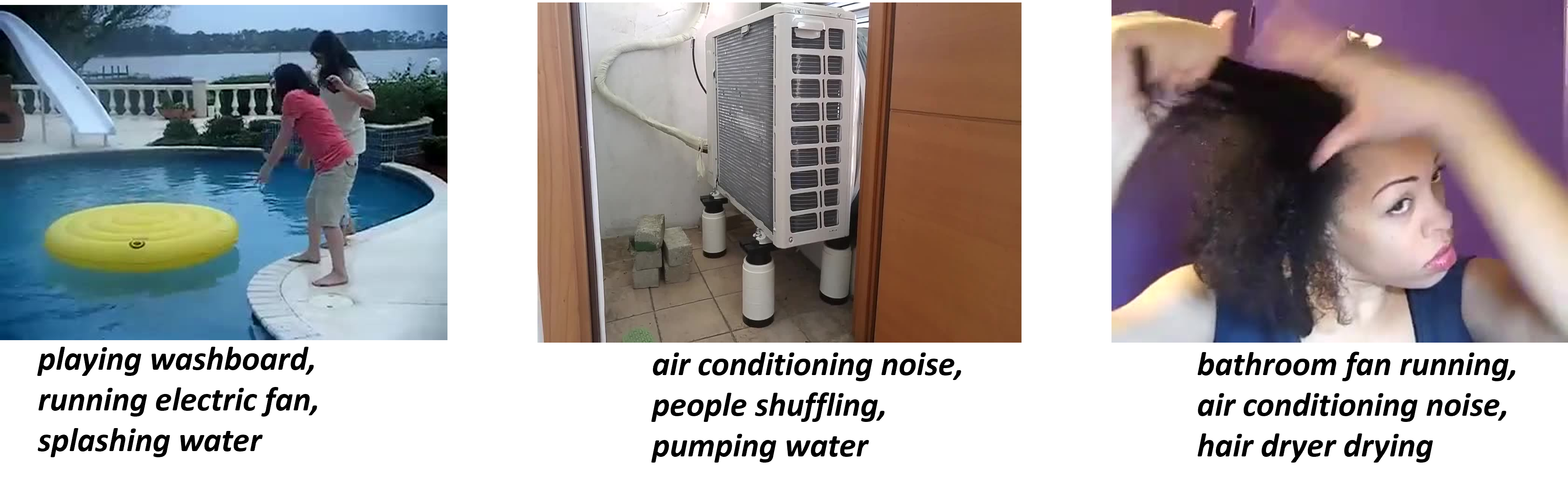}
		\caption{Attention model trained with BCE+margin loss (Model 3 in Table~\ref{tab:rec_vgg})}
		\label{fig:vgg_rec_att}
	\end{subfigure}
	\begin{subfigure}[t]{\textwidth}
		\centering
		\includegraphics[height=1.1in]{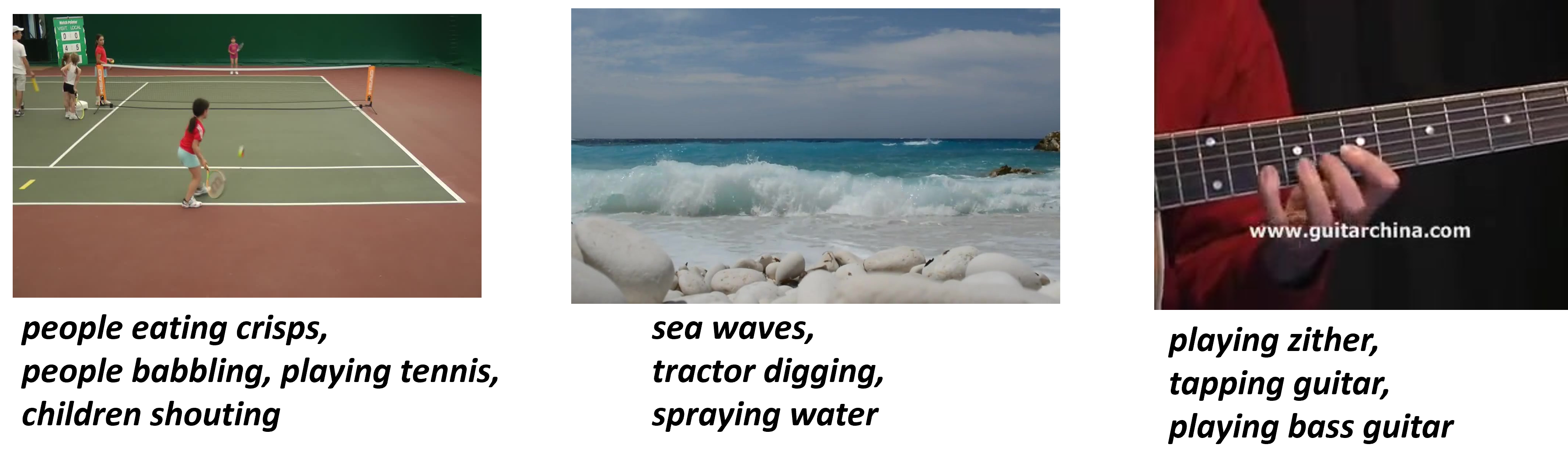}
		\caption{Attention model  trained with InfoNCE loss (Model 5 in Table~\ref{tab:rec_vgg}) }
		\label{fig:vgg_rec_contrast}
	\end{subfigure}
	\caption{Top sounds recommended by attention models for VGG-Sound videos} % (images best viewed in color)}	
\end{figure*}

\subsection{Audio-Visual Correlation Performance}   
%\vspace{-0.05in}
We  evaluated the  A-V correlation performance of the models on 3000 audio-visual pairs, each pair consisting of a video frame and 1-second audio sample extracted from the VGG-Sound test videos.  The test set consists of an equal number of correlated and uncorrelated pairs.  Table~\ref{tab:cor} shows the correlation accuracy, which measures the percentage of test pairs for which the audio-visual correlation was correctly classified  by the 3 approaches in Section~\ref{sec:models}. The first two models were trained to classify the audio-visual correlation. The attention model  (Section~\ref{sec:av_att})   trained with a combination of  binary cross-entropy (BCE) and margin-based contrastive loss (Equation \ref{eqn:margin} with margin=0.1)  improves the correlation accuracy by around 18\%, compared to the baseline model without attention. For the third model, the attention-based encoder was first trained with contrastive learning using InfoNCE loss to directly learn A-V representations (Section~\ref{sec:av_contrast}).  This pre-trained encoder was then combined with a binary classifier layer and the whole model was fine-tuned to classify the A-V correlation using BCE+margin-based contrastive loss. All 3 approaches classify the  correlation  based on the distance between the generated A-V embeddings. From Table~\ref{tab:cor}, we infer that the attention-based encoder is more effective in predicting the audio-visual correlation because it learns better audio and visual representations  that are closer when correlated and farther when they are uncorrelated.

\begin{figure*}[ht]
	\begin{subfigure}[t]{\textwidth}
		\centering
		\includegraphics[height=1.3in]{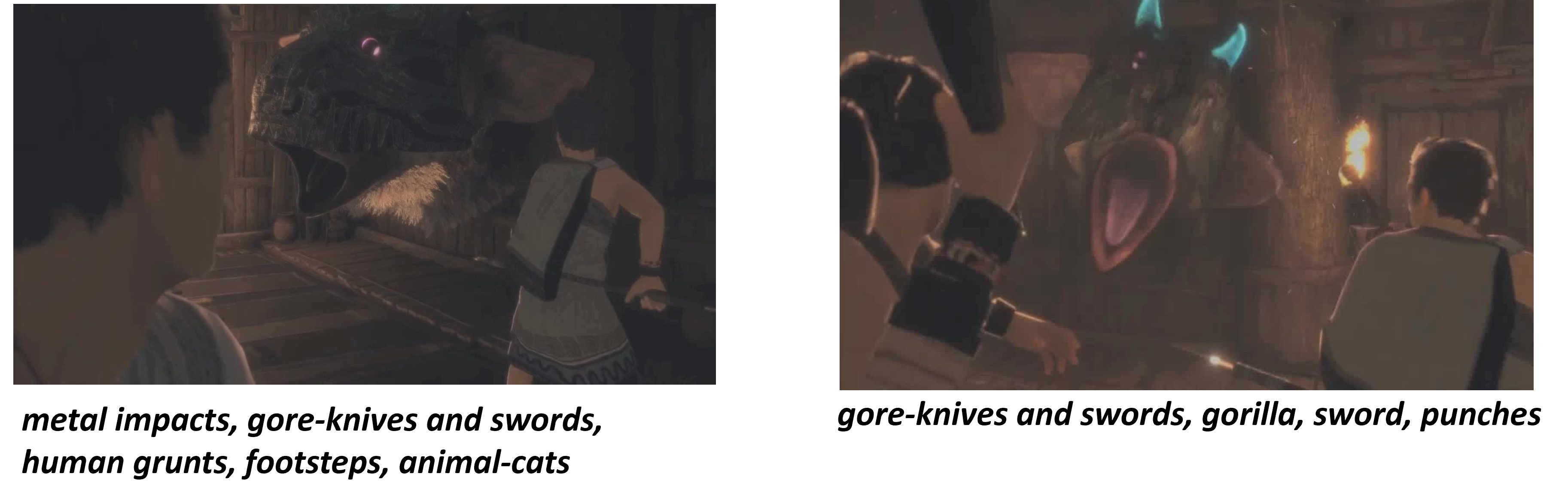}
		\caption{Recommendations with a good match (images best viewed in color).}
		\label{fig:game_good_rec_contrast}
	\end{subfigure}
    %\hfill
	\begin{subfigure}[t]{\textwidth}
		\centering
		\includegraphics[height=1.3in]{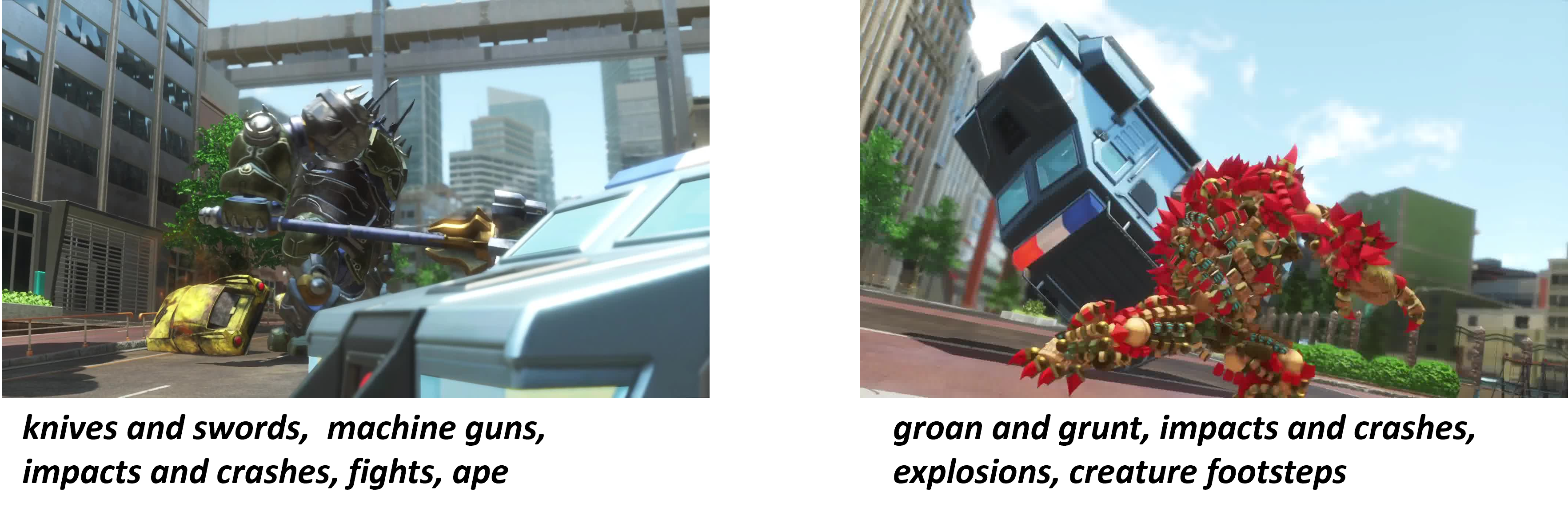}
		\caption{Recommendations with some mismatch (images best viewed in color).}
		\label{fig:game_poor_rec_contrast}
	\end{subfigure}
	\caption{Topmost sounds recommended for Gameplay videos by Model 5 in Table~\ref{tab:rec_vgg}. \\{\scriptsize{(The frames in (a) are from the video game The Last Guardian\texttrademark.  \copyright 2016 Sony Interactive Entertainment Inc. The Last Guardian is a trademark of Sony Interactive Entertainment America LLC. The frames in (b) are from Knack\texttrademark2.  \copyright2017 Sony Interactive Entertainment Inc. Knack is a trademark of Sony Interactive Entertainment LLC.)}}}
	\label{fig:game_rec}
\end{figure*}
%128 val data, 300 test data
%vgg: 58290:1280:3000
\subsection{Sound Recommendation Performance} 
We also evaluated the  A-V  representations learned   by the trained audio-visual models based on  the sound samples they recommended for the VGG-Sound and Gameplay test videos. The models recommended the top-k sound samples from an audio list or database. Each sample $i$ in the audio search list is labeled as  $(s_i, c_i)$, where $s_i$ is the sample label that uniquely identifies the audio sample, while $c_i$ is its category label, which may be shared by multiple audio samples. These labels are used to evaluate the top-k sample-level (or instance-level) and category-level recommendation accuracy, as follows. 
Let $R(v_j)=[S^R(v_j)=(s_1^R,\ldots,s_k^R); C^R(v_j)=(c_1^R,\ldots,c_k^R)]$ denote the top-k sounds recommended by a model for a video frame $v_j$, where $S^R(v_j)$ and $C^R(v_j)$ are the list of sample labels and category labels, respectively, for the top-k sounds recommended  for  $v_j$. Let $G(v_j)=[S^G(v_j)=(s_1^G,\ldots,s_n^G); C^G(v_j)=(c_1^G,\ldots,c_n^G)]$ denote the ground-truth audio samples for video frame $v_j$, where $S^G(v_j)$ and $C^G(v_j)$ are the list of sample labels and category labels for the ground-truth audio samples for  $v_j$. If $S^R(v_j) \cap S^G(v_j) \neq \phi$, there is a sample-level match and if $C^R(v_j) \cap C^G(v_j) \neq \phi$, there is a category-level match. Sample-level match is  stricter  than  category-level match. The top-k {\em{sample-level accuracy}} and top-k {\em{category-level accuracy}} of a model measure the percentage of test video frames for which there is a sample-level match and category-level match, respectively, between the top-k  sounds recommended by the model and the ground-truth.

{\bf{Recommendations for VGG-Sound Dataset:}} 
To evaluate the recommendation performance of a model on VGG-Sound, we randomly  extracted one video frame $v_i$ and its corresponding 1-second audio sample $a_i$, from each of the test videos. $a_i$ serves as the ground-truth for $v_i$ and is identified by $(s_i, c_i)$, where the sample label $s_i$ is the unique name of the video and $c_i$ is the category annotation of the video from  which $a_i$ is extracted. The audio samples $(a_1,\ldots,a_n)$ that are extracted from the test videos form the search list for sound recommendations.  We resized each extracted test video frame  to  224x224-dim and normalized it. We then generated the 128-dim visual embeddings for the test video frames  and  128-dim audio embedding for each of the samples in the audio search list, using the evaluation model  trained on the VGG-Sound training set. To recommend sounds for a video frame, we computed the Euclidean distance between its visual embedding and all of the audio embeddings in the search list and selected the top-k nearest audio samples. We then compared these top-k sound recommendations with the ground-truth to compute the top-k sample-level and category-level accuracy.% as described above. 

Table~\ref{tab:rec_vgg} shows the top-10 category-level and sample-level recommendation accuracy   for the VGG-Sound video frames. The first two models are trained with BCE classification loss and learn  audio-visual representations while  learning to classify the correlation.  The attention-based encoder model (Section~\ref{sec:av_att})  has a better performance than the baseline model  without attention (Section~\ref{sec:av_base}). When trained with a combination of  BCE and margin-based contrastive loss (Equation~\ref{eqn:margin} with $m=0.1$), the sample-level and category-level recommendation accuracy of the attention model (Model 3  in Table~\ref{tab:rec_vgg}) improves by around 10\% compared to the baseline model.  The margin loss term introduces an additional constraint to ensure that the correlated and uncorrelated pairs of audio-visual representations are separated by a margin of at least 0.1. Figure~\ref{fig:vgg_rec_att} shows the categories of the top sound samples recommended by Model 3 for some of the VGG-Sound test videos. In Models 4-7, contrastive learning is used to train the  attention-based encoder with different projection heads  and different loss functions (Section~\ref{sec:av_contrast}). The non-linear projector has 2 fully-connected layers with bias and a ReLU non-linearity in between, whereas the linear projector has only a single fully-connected layer with bias. For InfoNCE as well as NT-Xent loss, the model with linear projection  outperforms  the corresponding model with non-linear projection. Also, the model with linear projection trained with InfoNCE loss (Model 5) has  the best  recommendation accuracy and outperforms the linear model  trained with NT-Xent loss (Model 7). Figure~\ref{fig:vgg_rec_contrast} shows the labels of some of  the top sound samples recommended by the best performing model (Model 5) for some of the VGG-Sound video frames. 

We now summarize the recommendation results using VGG-Sound: 1) From Figure~\ref{fig:vgg_rec_att} and \ref{fig:vgg_rec_contrast}, we see that the attention-based models are able to recommend sound samples that correspond well with the visual scenes, which shows that these models  learn effective self-supervised representations and correspondence from audio-visual features extracted at different resolutions, even without explicit annotations, object detection, and sound separation. 2) Models trained with contrastive learning generate better  representations  and outperform the models that learn the representations using a classification objective for the correlation (Models 1-3). In particular, the best model (Model 5) improved the sample-level and category-level recommendation accuracy by 12\% and 14\%, respectively, compared to Model 3. We need to explore if cross-modal contrastive learning with data augmentation can further improve the performance.
%when learning to classify the A-V correlation

{\bf{Recommendations for Gameplay Dataset:}} 
We also evaluated the sound recommendation performance on 676 video frames extracted from 30 videos in the Gameplay test dataset. The sounds were recommended from a large internal audio database  consisting of more than 200K sound samples grouped into  4000 different categories.  We used the same procedure and model architectures for evaluation, as described above for the VGG-Sound dataset. Models 1-3 in Table~\ref{tab:rec_vgg} 
%that learn the audio and visual representations while learning to classify the A-V correlation  
were first trained on the Gameplay training dataset and then used to generate the visual embeddings for the test video frames and the audio embeddings for the more than 200K samples in the audio database (the audio samples in this database were not used during training). On the other hand, the contrastive learning  models  that were directly trained  to learn A-V representations (Models 4-7 in Table~\ref{tab:rec_vgg})  learn more generic representations. Hence,  to evaluate their generalizability, we used the contrastive learning models that were trained only on the VGG-Sound training set  to generate the embeddings for the Gameplay test video frames and audio  samples in the search database. Each model then used the  embeddings it generated to recommend the top-k samples from  the audio database for the 676 test scenes.

The Gameplay test videos do not have any  sample-level or category-level ground-truth for the sounds. 
%because it is hard to separate the audio mixed with the original video into its sources and match them with the samples in the audio database.  
Instead, a few sound designers manually recommended some sound samples from the audio database for our test videos.  While our models generate the top-k recommendations by comparing the visual embedding  with the audio embeddings of all of the  audio samples in  the database, the manual recommendations are not based on an exhaustive search of the database. So it is hard to get an exact match between the manual  and model recommendations. Nevertheless,  the manual recommendations serve as approximate ground-truth to evaluate the relative performance of our models. The sample-level recommendation accuracy of the attention-based  model was 10x higher than that of the baseline model. Both of these models were trained with the Gameplay dataset to classify the A-V correlation. The attention model   trained with InfoNCE loss (Model 5 in Table~\ref{tab:rec_vgg}) to directly learn the representations, further improved the sample-level and category-level accuracy of the Gameplay recommendations by 7\% and 24\%, respectively, even though this model was trained on VGG-Sound and there was a significant domain gap between the VGG-Sound and Gameplay dataset.  Figure~\ref{fig:game_rec} shows the categories of the top sound samples recommended by Model 5 for some of the Gameplay test videos. Figure~\ref{fig:game_good_rec_contrast} shows video frames for which the model generated recommendations  were a good match for the visual scene, while Figure~\ref{fig:game_poor_rec_contrast} are examples with some mismatch. These results show that the attention-based encoder  is able to learn effective audio and visual representations based on features extracted at different resolutions, even from complex datasets that have multiple visual objects and audio mixed with multiple sound sources. Furthermore, the attention model trained  with cross-modal contrastive learning is able to improve the recommendation performance on the Gameplay dataset, even though the model was  trained on the less noisy VGG-Sound dataset. This shows the generalizability of the cross-modal contrastive learning approach.  As future work, we plan to improve the Gameplay recommendations  further by using a contrastive learning model trained on the Gameplay dataset.
\section{Conclusions}
\label{sec:conc}

We presented self-supervised approaches that combine a novel attention-based encoder with a projection module  to learn audio and visual representations based on their correspondence, by training on  unlabeled videos. The attention encoder  extracts  local and global audio-visual features at different resolutions and  attends to them based on their relative importance. This helps in disentangling latent features extracted from videos having multiple objects and sound sources, without using explicit object detection or audio source separation.
The model projects correlated audio and visual inputs closer in the latent space compared to uncorrelated audio-visual pairs. Our results show that when trained with  cross-modal contrastive learning,  the attention encoder is able to learn better audio and visual representations even without data augmentation and significantly improve the sound recommendation accuracy on video datasets with different complexities. The generated representations  can be used for unimodal and multi-modal similarity and  multimodal content creation tasks, such as creative video editing using recommended sound effects based on the objects and  context in the visual scenes. \\
{\small{
{\bf{Acknowledgments:}} The author would like to thank Norihiro Nagai, Isamu Terasaka, Kiyoto Shibuya, and Saket Kumar for the feedback and support they provided.
}}

{\small
	\bibliographystyle{splncs04}
	\bibliography{vis_sound}
}

\end{document}